%% file: main_arXiv.tex
\documentclass{article}

\usepackage[utf8]{inputenc} % allow utf-8 input
\usepackage[T1]{fontenc}    % use 8-bit T1 fonts
\usepackage{hyperref}       % hyperlinks
\usepackage{url}            % simple URL typesetting
\usepackage{booktabs}       % professional-quality tables
\usepackage{amsthm}
\usepackage{amsfonts}       % blackboard math symbols
\usepackage{nicefrac}       % compact symbols for 1/2, etc.
\usepackage{microtype}      % microtypography
\usepackage{authblk}
\usepackage{fullpage}
\usepackage{xspace}
\usepackage{mathtools}
\usepackage{algorithm}
\usepackage{algorithmic}
\usepackage{natbib}
\usepackage{subfigure}

\def\arxiv{1}

\input{macros}

\title{Hiding in the Crowd: A Massively Distributed Algorithm for Private Averaging with Malicious Adversaries}

\author[1]{Pierre Dellenbach}
\author[1]{Aurélien Bellet\thanks{Corresponding author: \texttt{first.last@inria.fr}}}
\author[1]{Jan Ramon}
\affil[1]{INRIA}
\date{}

\begin{document}

\maketitle

\input{subfiles/abstract.tex}

\input{subfiles/intro.tex}

\input{subfiles/setting.tex}

\input{subfiles/related_work.tex}

\input{subfiles/gopa.tex}

\input{subfiles/verif.tex}

\input{subfiles/experiments.tex}

\input{subfiles/conclu.tex}

\paragraph{Acknowledgments} This research was partially supported by grant ANR-16-CE23-0016-01 and by a grant from CPER Nord-Pas de Calais/FEDER DATA Advanced data science and technologies 2015-2020. The work was also partially supported by ERC-PoC SOM 713626.

\bibliographystyle{apalike}
\bibliography{main}

\clearpage
\appendix
\renewcommand{\thesection}{Appendix~\Alph{section}}
\renewcommand{\thesubsection}{\Alph{section}.\arabic{subsection}}

\section*{SUPPLEMENTARY MATERIAL}
\input{subfiles/supplementary.tex}

\end{document}

%% file: macros.tex
\DeclarePairedDelimiter{\ceil}{\lceil}{\rceil}

\DeclareMathOperator{\lcm}{lcm}

\DeclareMathOperator{\diag}{diag}

\def \ZN2Z{\mathbb{Z}_{N^2}^*}
\def \ZNZ{\mathbb{Z}_{N}^*}

\newcommand{\userset}{U}
\newcommand{\avgx}{X^{avg}}
\newcommand{\avgxhat}{{\hat{X}}^{avg}}

  \newcommand{\usign}[2]{\xi({#1},{#2})}
  \newcommand{\smalluser}[2]{\zeta({#1},{#2})}

\newcommand{\gopa}{\textsc{Gopa}\xspace}

\newcommand{\sigmaX}{\sigma_X}
\newcommand{\sigmaDelta}{\sigma_{\delta}}
\newcommand{\edgeset}{E}
\newcommand{\neighborset}{N}
\newcommand{\honestuserset}{\userset^{H}}
\newcommand{\honestedgeset}{\edgeset^{H}}
\newcommand{\honestordedgeset}{\edgeset^{H}_{<}}
\newcommand{\honestneighborset}{\neighborset^{H}}

\newcommand{\allrvs}{\mathcal{V}}
\newcommand{\rvunitvect}[1]{e_{{#1}}}

\newtheorem{baseline}{Baseline}
\newtheorem{assumption}{Assumption}

\newtheorem{proposition}{Proposition}
\newtheorem{remark}{Remark}
\newtheorem{theorem}{Theorem}

%% file: subfiles/abstract.tex
% !TEX root = ../main_supp.tex

\begin{abstract}
  The amount of personal data collected in our everyday interactions with connected devices offers great opportunities for innovative services fueled by machine learning, as well as raises serious concerns for the privacy of individuals.
  In this paper, we propose a massively distributed protocol for a large set of users to privately compute averages over their joint data, which can then be used to learn predictive models. Our protocol can find a solution of arbitrary accuracy, does not rely on a third party and preserves the privacy of users throughout the execution in both the honest-but-curious and malicious adversary models. Specifically, we prove that the information observed by the adversary (the set of maliciours users) does not significantly reduce the uncertainty in its prediction of private values compared to its prior belief. The level of privacy protection depends on a quantity related to the Laplacian matrix of the network graph and generally improves with the size of the graph.
  Furthermore, we design a verification procedure which offers protection against malicious users joining the service with the goal of manipulating the outcome of the algorithm.
% \keywords{\aurelien{TODO \and TODO \and TODO}}
\end{abstract}

%% file: subfiles/intro.tex
% !TEX root = ../main_supp.tex

\section{Introduction}
\label{sec:intro}

Through browsing the web, engaging in online social networks and interacting with connected devices, we are producing ever growing amounts of sensitive personal data. This has fueled the massive development of innovative personalized services which extract value from users' data using machine learning techniques. In today's dominant approach, users hand over their personal data to the service provider, who stores everything on centralized or tightly coupled systems hosted in data centers. Unfortunately, this poses important risks regarding the privacy of users. %For instance, it is difficult for users to control how their data is used, and data concentration makes the system vulnerable to large-scale cyber-attacks.
To mitigate these risks, some approaches have been proposed to learn from datasets owned by several parties who do not want to disclose their data. However, they typically suffer from some drawbacks: (partially) homomorphic encryption schemes \citep{paillier1999public,Graepel2012a,Aslett2015a} require the existence of a trusted third party, secure multi-party computation techniques \citep{Yao,Lindell2009a} are generally intractable when the number of parties is large, and exchanging noisy sketches of the data through (local) differential privacy \citep{Dwork2006a,Duchi2012b} only provides approximate solutions which are quite inaccurate in the highly distributed setting considered here. Furthermore, many of these techniques are not robust to the presence of malicious parties who may try to manipulate the outcome of the algorithm.

In this paper, our goal is to design a massively distributed protocol to collaboratively compute averages over the data of thousands to millions of users (some of them honest-but-curious and some corrupted by a malicious party), with arbitrary accuracy and in a way that preserves their privacy. For machine learning algorithms whose sufficient statistics are averages (e.g., kernel-based algorithms in primal space and decision trees), this could be used as a primitive to privately learn more complex models.
The approach we propose is fully decentralized: users keep their own data locally and exchange information asynchronously over a peer-to-peer network (represented as a connected graph), without relying on any third party. Our algorithm (called \gopa: GOssip for Private Averaging) draws inspiration from a randomized gossip protocol for averaging \citep{random_gossip}, augmented with a first phase involving pairwise exchanges of noise terms so as to mask the private values of users without affecting the global average.
We first analyze the correctness of the algorithm, showing that the addition of noise has a mild effect on the convergence rate.
We then study the privacy guarantees of \gopa in a Bayesian framework, where the adversary has some prior belief about the private values. Specifically, we give an exact expression for the posterior variance of the adversary after he has observed all the information output by the protocol, and show that the variance loss is negligible compared to the prior. This is equivalent to showing that the uncertainty in the adversary's predictions of the private values has not been significantly reduced. Interestingly, the proportion of preserved variance depends on the variance of the noise used to mask the values but also on an interpretable quantity related to the Laplacian matrix of the network graph.
% We then carefully study the privacy guarantees of \gopa in the context of Pufferfish privacy \citep{Kifer2014a}, a recently proposed framework to design privacy definitions. Specifically, we define a relaxed notion of Differential Privacy \citep{Dwork2006a} which assumes that the adversary has some prior uncertainty about the private value of each honest user (instead of potentially knowing all private values except for one honest user).
% We prove that \gopa satisfies this definition with a privacy parameter which depends on the variance of the noise used to mask the values as well as on an interpretable quantity related to the Laplacian matrix of the network graph.
To the best of our knowledge, we are the first to draw a link between privacy and a graph smoothing operator popular in semi-supervised learning \citep{Zhu2002a,Zhou2003a}, multi-task learning \cite{Evgeniou2004a} and signal processing \citep{Shuman2013a}. We show how this result motivates the use of a random graph model to construct the network graph so as to guarantee strong privacy even under rather large proportions of malicious users, as long as the number of users is big enough. The practical behavior of \gopa is illustrated on some numerical simulations.
Finally, we further enhance our protocol with a verification procedure where users are asked to publish some values in encrypted form, so that cheaters trying to manipulate the output of the algorithm can be detected with high probability while preserving the aforementioned privacy guarantees.

% \aurelien{maybe stress more the idea that we turn the massively distributed setting into an advantage by diluting the knowledge across many parties (``not all eggs in the same basket'')}

% In this paper, we take an alternative approach. We propose a fully decentralized (peer-to-peer) protocol which allows thousands to millions of users to collaboratively compute averages over their joint data with arbitrary accuracy and in a way that preserves their privacy. For machine learning algorithms whose sufficient statistics are averages (e.g., kernel-based algorithms in primal space and decision trees), the proposed method can be used as a primitive to privately learn more complex models. Our algorithm draws inspiration from a randomized gossip protocol for decentralized averaging \citep{random_gossip}, augmented with a first phase involving the addition of noise terms so as to mask the true values of users' data. We further enhance our protocol with a verification procedure where users are asked to publish some values in encrypted form, so that cheaters trying to manipulate the output of the algorithm can be detected with high probability while preserving privacy guarantees.

The rest of this paper is organized as follows. Section~\ref{sec:setting} describes the problem setting, including our adversary and privacy models.
Section~\ref{sec:related} presents some background on (private) decentralized averaging and partially homomorphic encryption, along with related work and baseline approaches.
Section~\ref{sec:gopa} introduces the \gopa algorithm and studies its convergence qs well as privacy guarantees. Section~\ref{sec:verif} describes our verification procedure to detect cheaters. Finally, Section~\ref{sec:exp} displays some numerical simulations. Proofs can be found in the supplementary material.

%% file: subfiles/setting.tex
\section{Preliminaries}
\label{sec:setting}

We consider a set $\userset=\{1,\dots,n\}$ of $n\geq 3$ users. Each user $u\in U$ holds a personal value $X_u\in\mathbb{R}$, %, which is the realization of a random variable with probability distribution $P_u$ (independent of other users).
which can be thought of as the output of some function applied to the personal data of $u$ (e.g., a feature vector describing $u$). The users want to collaboratively compute the average value $\avgx=\frac{1}{n}\sum_{u=1}^n X_u$ while keeping their personal value private. Such a protocol could serve as a building block for privately running machine learning algorithms which interact with the data only through averages, such as linear regression models (ordinary least-squares, ridge regression), decision trees and gradient descent for empirical risk minimization problems.
%We assume that the personal values lie in some bounded range, namely $X^{min} \le X_u \le X^{max}$ for all $u\in U$, and 
We denote by $X$ the vector $X=[X_1,\dots,X_n]^\top\in\mathbb{R}^n$.

Instead of relying on a central server or any third party to exchange information, users communicate over a peer-to-peer network represented by a connected undirected graph $G=(U,E)$, where $(u,v)\in E$ indicates that users $u$ and $v$ are neighbors in $G$ and can exchange messages directly. For a given user $u$, we denote by $N(u)=\{v : (u,v)\in E\}$ the set of its neighbors. We denote by $A$ the adjacency matrix of $G$, by $d=(d_1,\dots,d_n)$ the degree vector ($d_u=\sum_v A_{uv})$ and by $L=\diag(d)-A$ its Laplacian matrix.

\subsection{Adversary Models}

We consider two commonly adopted adversary models for users, which were formalized
by \citet{goldreich1998secure} and are used in the design of
many secure protocols.
An \emph{honest-but-curious} (\emph{honest} for short) user will follow
the protocol specification, but can use all the information received
during the execution to infer information about other users.
In contrast, a \emph{malicious user} may deviate from the protocol
execution by sending incorrect values at any point (but we assume that they follow the required communication policy; if not, this can be easily detected). Malicious users can collude, and thus will be seen as a single malicious party who has access to all information collected by malicious users.
We only restrict the power of attackers by requiring that 
honest users communicate through secure channels, which means that malicious users only observe information during communications
they take part in.

Each user in the network is either honest or malicious, and honest users do not know whether other nodes are honest or malicious. We denote by $U^H\subseteq U$ the set of honest users and by $f=1-|U^H|/n$ the proportion of malicious users in the network. We also denote by $G^H = (U^H, E^H)$ the subgraph of $G$ induced by $U^H$, so that $E^H = \{(u,v) \in E : u,v \in U^H\}$. Throughout the paper, we will rely on the following natural assumption on $G^H$ (we will discuss how to generate $G$ such that it holds in practice in Section \ref{sec:graphs}).
\begin{assumption}
\label{assump:GH}
The graph of honest users $G^H$ is connected.
\end{assumption}
This implies that there exists a path between any two honest users in the full graph $G$ which does not go through a malicious node.
In the rest of the paper, we will use the term \emph{adversary} to refer to the set of malicious users (privacy with respect to a honest user can be obtained as a special case).

\subsection{Privacy Model}

Recall that our goal is to design a protocol which deterministically outputs the \emph{exact} average (which we argue does not reveal much information about individual values in the large-scale setting we consider). This requirement automatically rules out Differential Privacy \citep{Dwork2006a} as the latter implies that the output of the protocol has to be randomized.

We take a Bayesian, semantic view of privacy promoted by several recent papers \citep{Kasiviswanathan2014a,Li2013a,Kifer2014a,He2014a}. We consider a family of prior distributions which represent the potential background knowledge that the adversary may have about the private values of honest users (since the adversary controls the malicious users, we consider he has exact knowledge of their values). We will denote by $P(X_u)$ the prior belief of the adversary about the private value of some honest user $u\in U^H$. Given all the information $\mathcal{I}$ gathered by the adversary during the execution of the protocol, the privacy notion we consider is that the ratio of prior and posterior variance of the private value $X_u$ is lower bounded by $1-\epsilon$ for some $\epsilon\in[0,1]$:
\begin{equation}
\label{eq:privacy-def}
\frac{var(X_u \mid \mathcal{I})}{var(X_u)} \geq 1-\epsilon
\end{equation}
The case $\epsilon=1$ describes the extreme case where observing $\mathcal{I}$ removed all uncertainty about $X_u$. On the other hand, when $\epsilon=0$, all variance was preserved.

It is important to note that \eqref{eq:privacy-def} with $\epsilon$ close to $0$ does not guarantee that the adversary learns almost nothing from the output $\mathcal{I}$: in particular, the posterior expectation $\mathbb{E}[X_u\mid \mathcal{I}]$ can largely differ from the prior expectation $\mathbb{E}[X_u]$, especially if the observed global average was very unlikely under the prior belief. %\footnote{Consider for instance the case where the user values are positive with high probability under the prior belief of the adversary, yet the observed global average is negative.}
This is related to the ``no free lunch'' theorem in data privacy \citep{Kifer2011a}, see also the discussions in \cite{Kasiviswanathan2014a,Li2013a}.
What \eqref{eq:privacy-def} does guarantee, however, is that the squared error that the adversary expects to make by predicting the value of some private $X_u$ after observing $\mathcal{I}$ is almost the same as it was before observing $\mathcal{I}$. In other words, the \emph{uncertainty} in its prediction has not been significantly reduced by the participation to the protocol. This is formalized by the following remark.

\begin{remark}[Expected squared error]
Assume that \eqref{eq:privacy-def} is satisfied. Let $Y_u=\mathbb{E}[X_u]$ be the best prediction (in terms of expected square error) that the adversary can make given its prior belief. After observing the output $\mathcal{I}$, the adversary can make a new best guess $Y_u^{\mathcal{I}}=\mathbb{E}[X_u\mid \mathcal{I}]$. We have:
\begin{equation*}
\frac{\mathbb{E}[(Y_u^\mathcal{I}-X_u)^2]}{\mathbb{E}[(Y_u-X_u)^2]} = \frac{var(X_u\mid \mathcal{I})}{var(X_u)}=1-\epsilon.
\end{equation*}
\end{remark}

Our results will be valid for Gaussian prior distributions of the form $X_u\sim\mathcal{N}(0, \sigma_X^2)$, for all $\sigma_X^2>0$.\footnote{We use Gaussian distributions for technical reasons. We expect similar results to hold for other families of distributions which behave nicely under conditioning and linear transformations, such as the exponential family.} We assume for simplicity that the prior variance is the same for all $u$, but our analysis straightforwardly extends to the more general case where $X_u\sim\mathcal{N}(0, \sigma_{X_u}^2)$. We use centered Gaussians without loss of generality, since \eqref{eq:privacy-def} depends only on the variance.

\begin{remark}[Privacy axioms]
One can show that our notion of privacy \eqref{eq:privacy-def} satisfies the axioms that any reliable privacy definition should satisfy according to \citet{Kifer2012a}, namely ``transformation invariance'' and ``convexity''.
\end{remark}

%% file: subfiles/related_work.tex
\section{Background and Related Work}
\label{sec:related}

\subsection{Decentralized Averaging}

\begin{algorithm}[t]
\floatname{algorithm}{Algorithm}
  \caption{Randomized gossip \citep{random_gossip}}
  \begin{algorithmic}[1]
    \STATE{{\textbf{Input:}} graph $G=(\userset, E)$, initial values $(X_u(0))_{u\in \userset}$}
    \FOR{$t\in\{1,2,...\}$} 
    \STATE{Draw $(u,v)$ uniformly at random from $E$}
    \STATE{Set $X_u(t), X_v(t) \leftarrow \frac{1}{2}(X_u(t-1) + X_v(t-1))$}
    \ENDFOR
  \end{algorithmic}
  \label{alg:gossip}
\end{algorithm}

The problem of computing the average value $X^{avg}$ of a set of users in a fully decentralized network without a central coordinator has been extensively studied \citep[see for instance][]{Tsitsiklis1984a,Kempe2003a,random_gossip}. In most existing approaches, users iteratively compute the weighted average between their value and those of (a subset of) their neighbors in the network. We focus here on the randomized gossip algorithm proposed by \citet{random_gossip} as it is simple, asynchronous and exhibits fast convergence. Each user has a clock ticking at the times of a rate 1 Poisson process. When the clock of an user $u$ ticks, it chooses a random neighbor $v$ and they average their current value. As discussed in \citet{random_gossip}, one can equivalently consider a single clock ticking at the times of a rate $n$ Poisson process and a random edge $(u,v)\in E$ drawn at each iteration. The procedure is shown in Algorithm~\ref{alg:gossip}, and one can show that all users converge to $X^{avg}$ at a geometric rate.

Although there is no central server collecting all users' data, the above gossip algorithm is not private since users must share their inputs directly with others. A way to ensure privacy is that each user locally perturbs its own input before starting the algorithm so as to satisfy local differential privacy \citep{Duchi2012b,Kairouz2016a}.

\begin{baseline}
\label{base:gossip_localDP}
Assume $|X_u|\leq B_X$ for some finite $B_X\in\mathbb{R}$. Each user $u$ computes a perturbed value $\widetilde{X}_u=X_u+\eta_u$, where $\eta_u$ is a noise value with Laplacian distribution $P(\eta_u)= \exp(-|\eta_u|/b)/2b$ with $b=2B/\epsilon$. This guarantees that $\widetilde{X}=[\widetilde{X}_1,\dots,\widetilde{X}_n]^T$ is an $\epsilon$-differentially private approximation of $X$ \citep[see e.g.,][]{Dwork2008a}, hence running Algorithm~\ref{alg:gossip} on $\widetilde{X}$ is also $\epsilon$-differentially private.
\end{baseline}

Unfortunately, this protocol converges to an approximate average $\avgxhat=\sum_{u=1}^n (X_u+\eta_u)/n$, and the associated RMSE
$\sqrt{\mathbb{E}[(\avgx-\avgxhat)^2]}=b\sqrt{2/n}$
can be high even for a large number of users. For instance, for $n=10^4$ users, $\epsilon=0.1$ and $B=0.5$, the RMSE is approximately $0.14$.

Other attempts have been made at designing privacy-preserving decentralized averaging protocols under various privacy and attack models, and additional assumptions on the network topology. For an adversary able to observe all communications in the network, \citet{Huang2012a} proposed an $\epsilon$-differentially private protocol where users add exponentially decaying Laplacian noise to their value. The protocol converges with probability $p$ to a solution with an error radius of order $O(1/\epsilon\sqrt{(1-p)n})$. \citet{Manitara2013a} instead proposed that each user iteratively adds a finite sequence of noise terms which sum to zero, so that convergence to the exact average can be achieved. The adversary consists of a set of curious nodes which follow the protocol but can share information between themselves. Under some assumption on the position of the curious nodes in the network, their results prevent {\it perfect} recovery of private values but do not bound the estimation accuracy that the adversary can achieve. \citet{Mo2014a} proposed to add and subtract exponentially decaying Gaussian noise, also ensuring convergence to the exact average. They consider only honest-but-curious nodes and rely on a strong topological assumption, namely that there is no node whose neighborhood includes another node and its entire neighborhood. Their privacy guarantees are in the form of lower bounds on the covariance matrix of the maximum likelihood estimate of the private inputs that any node can make, which are difficult to interpret and very loose in practice.
Finally, \citet{Hanzely2017a} introduced variants of Algorithm~\ref{alg:gossip} which intuitively leak less information about the private inputs, but do not necessarily converge to the exact average. Importantly, they do prove any formal privacy guarantee.

In the above approaches, privacy is typically achieved by asking each user to independently add decaying local noise to its private value, which results in accuracy loss and/or weak privacy guarantees.
In contrast, the protocol we propose in this paper relies on {\it sharing} zero-sum noise terms across users so as to ``dilute'' the knowledge of private values into the crowd. This will allow a flexible trade-off between privacy and convergence rate even when a node has a large proportion of malicious neighbors, while ensuring the convergence to the exact average. Furthermore and unlike all the above approaches, we provide a verification procedure to prevent malicious nodes from manipulating the output of the algorithm. This verification procedure relies on partially homomorphic encryption, which we briefly present below.

\subsection{Partially Homomorphic Encryption}
\label{sec:paillier}

A standard technique for secure computation is to rely on a partially homomorphic encryption scheme to cipher the values while allowing certain operations to be carried out directly on the cypher text (without decrypting).

We use the popular Paillier cryptosystem \citep{paillier1999public}, which is additive.
Formally, one generates a public (encryption) key $K^{pub}=(N,g)$, where $N=pq$ for $p\neq q$ two independent large primes and $g\in\mathbb{Z}^*_{N^2}$, and a secret (decryption) key $K^{priv} = \lcm(p-1,q-1)$. A message $m\in\mathbb{Z}^*_N$ can then be encrypted into a cypher text with
\begin{equation}
    \label{eq:paillier_enc}
    \varepsilon(m) = g^m \cdot r^N \bmod N^2,
\end{equation}
where $r$ is drawn randomly from $\mathbb{Z}^*_N$. With
knowledge of the secret key, one can recover $m$ from $\varepsilon(m)$ based on the fact that
$\frac{\varepsilon(m)^{\lambda}-1}{g^{\lambda}-1} \equiv m \bmod N$.
Denote the decryption operation by $\varepsilon^{-1}$. Paillier satisfy the following homomorphic property for any
$m_1, m_2 \in \mathbb{Z}^*_N$:
\begin{equation}
\label{eq:paillier_hom}
\varepsilon^{-1}(\varepsilon(m_1) \cdot \varepsilon(m_2) \bmod N^2) = m_1 + m_2 \bmod N,
\end{equation}
hence $\varepsilon(m_1) \cdot \varepsilon(m_2)$ is a valid encryption of $m_1+m_2$.
For the purpose of this paper, we will
consider the Paillier encryption scheme as perfectly secure (i.e., the computational complexity needed to break the encryption is beyond reach of any party).
We can use the Paillier scheme to design a second simple baseline for private averaging.

\begin{baseline}
\label{base:paillier}
Consider $X_1,\dots,X_n\in\mathbb{Z}^*_N$ and assume that users trust two central honest-but-curious entities: the server and the third party. The server generates a Paillier encryption scheme $(K^{pub},K^{priv})$ and broadcasts $K^{pub}$ to the set of users. Each user $u$ computes
$\varepsilon(X_u)$ and sends it to the third party. Following
\eqref{eq:paillier_hom}, the third party then computes
$\prod_{u=1}^n\varepsilon(X_u)$ and
sends it to the server, which can obtain $X^{avg}$ by decrypting the message (and dividing by $n$). 
\end{baseline}
If the server and the third party are indeed honest, nobody observes any useful information except the outcome $\avgx$. But if they are not honest, various breaches can
occur. For instance, if the third party is malicious, it can send an incorrect output to the server. The third party could also send the encrypted values of some users to the server, which can decrypt them using the private key $K^{priv}$. Similarly, if the server is malicious, it can send $K^{priv}$ to the third party
which can then decrypt all of the users' private values.

In contrast, we will design a protocol which eliminates the need for such central trusted entities and instead distributes the trust across many users in the network (Section~\ref{sec:gopa}). We will however rely on homomorphic encryption to detect potential malicious users (Section~\ref{sec:verif}).

%% file: subfiles/gopa.tex
\section{GOPA: Private Gossip Averaging Protocol}
\label{sec:gopa}

\subsection{Protocol Description}
\label{sec:protocol}

We describe our \gopa protocol (GOssip for Private Averaging), which works in two phases.
In the first phase (\emph{randomization phase}), users mask their private value by adding noise terms that are correlated with their neighbors, so that the global average remains unchanged. In the second phase (\emph{averaging phase}), users average their noisy values.
For simplicity of explanation, we abstract away the communication mechanisms, i.e., contacting a neighbor and exchanging noise are considered as atomic operations.

\begin{algorithm}[t]
\floatname{algorithm}{Algorithm}
  \caption{Randomization phase of \gopa}
  \begin{algorithmic}[1]
    \STATE{\textbf{Input:}} graph $G=(\userset, E)$, private values $(X_u)_{u\in \userset}$, distribution $\mu:\mathbb{R} \to [0,1]$
    \FORALL{neighbor pairs $(u,v)\in E$ s.t. $u<v$}
    \STATE{$u$ and $v$ jointly draw a random number $\delta\in\mathbb{R}$ from $\mu$}
    \STATE{$\delta_{u,v}\gets \delta$, $\delta_{v,u}\gets -\delta$}
    \ENDFOR
    \FORALL{users $u\in\userset$}
    \STATE{$\Delta_u\gets \sum_{v\in N(u)} \delta_{u,v}$, $\widetilde{X}_u\gets X_u+\Delta_u$}
    \ENDFOR
    \STATE{\textbf{Output:}} noisy values $(\widetilde{X}_u)_{u\in \userset}$
  \end{algorithmic}
  \label{alg:rndphase}
\end{algorithm}

\textbf{Randomization phase.} Algorithm \ref{alg:rndphase} describes the first phase of \gopa, during which all neighboring nodes $(u,v)\in E$ contact each other to exchange noise values. Specifically, they jointly draw a random real number $\delta$ from a probability distribution $\mu$ (a parameter of the algorithm that the community agrees on), that $u$ will add to its private value and $v$ will subtract. Following the common saying ``don't put all your eggs in one basket'', each user thereby distributes the noise masking his private value across several other users (his direct neighbors but also beyond by transitivity), which will provide some robustness to malicious parties.
The idea is reminiscent of one-time pads \citep[see for instance][Section 3 therein]{Bonawitz2017a}, but our subsequent analysis will highlight the key role of the network topology and show that we can tune the magnitude of the noise so as to trade-off between privacy on the one hand, and convergence speed as well as the impact of user drop out on the other hand.
The result of this randomization phase is a set of noisy values $\widetilde{X}=[\widetilde{X}_1,\dots,\widetilde{X}_n]^\top$ for each user, with the same average value $\avgx$ as the private values. Note that each user $u$ exchanges noise exactly once with each of his neighbors, hence the noisy value $\widetilde{X}_u$ consists of $d_u$ noise terms.

% Every user $u$ collects all exchanged noise, sums it, and at the end adds it to its own private value $X_u$ to obtain a noisy version $\widetilde{X}_u$.
% % Line \ref{ln:selectNeighbor} of the algorithm requires that users select neighbors who are not overloaded by noise exchange requests.  This is not strictly needed for the correctness of the algorithm but ensures a good convergence rate.
% The individual noise values are drawn from a probability distribution the community agrees on (a parameter $\mu$ of the algorithm).
%, and is bounded in absolute value by $B$.
% For simplicity of explanation, we abstract away the communication mechanisms, e.g. selecting a neighbor and exchanging noise are considered as atomic operations.

% \aurelien{For randomization phase, in light of the new form of the privacy results, I think we should make the connections deterministic: contact each neighbor a fixed number of times to exchange noise (actually, only 1 time since exchanging more times is equivalent in this case to increasing noise variance). This would simplify things: (i) no need for $R, R^{in}, R^{out}$, (ii) it can happen at the same time as network construction, (iii) easier interpretation of privacy result (the Laplacian is simply the Laplacian of the subgraph of honest users), (iv) if we use the topological assumption it is enough to ensure connected graph and hence smoothing over all honest users.}

\textbf{Averaging phase.} In the second phase, the users start from their noisy values $\widetilde{X}$ obtained in the randomization phase and simply run the standard randomized gossip averaging algorithm (Algorithm~\ref{alg:gossip}).

\subsection{Correctness and Convergence of \gopa}
\label{sec:conv}

  % The algorithm cannot converge before the first phase is
  % completed due to the added noise. However, after all noise has been
  % added and has canceled out, convergence to the desired average can
  % be reached, as shown by the following result.

  In this section, we study the correctness and convergence rate of \gopa. Let $\widetilde{X}(t)=[\widetilde{X}_1(t),\dots,\widetilde{X}_n(t)]^\top$ be the values after $t\geq 0$ iterations of the averaging phase (Algorithm~\ref{alg:gossip}) initialized with the noisy values $\widetilde{X}$ generated by the randomization phrase (Algorithm~\ref{alg:rndphase}). Following the seminal work of \citet{random_gossip}, we will measure the convergence rate in terms of the \emph{$\tau$-averaging time}. Given $\tau\in(0,1)$, the $\tau$-averaging time is the number of iterations $t(\tau)$ needed to guarantee that for any $t\geq t(\tau)$:
  \begin{equation}
  \label{eq:tave}
  \Pr(\|\widetilde{X}(t)-X^{avg}\mathbf{1}\|/\|X\| \geq \tau) \leq \tau.
  \end{equation}
  Note that the error in \eqref{eq:tave} is taken relatively to the original set of values $X$ to account for the impact of the addition of noise on the convergence rate. We have the following result for the case where all users are honest (we will lift this requirement in Section~\ref{sec:verif}).

  % Note that before the first phase is completed the system cannot
  % converge to the average because of the noise added. So it is
  % important to make sure that the first phase terminates. We will
  % not enter the specifics, however, it is possible to do it. One can
  % simply ask that each user finish phase 1 in less than $T_1$
  % miliseconds (and start some communications with his neighbours
  % when time starts to run out). In the \textit{HBC} model, this is a
  % good enough insurance, and if some users are malicious, we simply
  % need to add the publication of the absolute time of communication
  % in the verification step.

  \begin{proposition}[Correctness and convergence rate]
    \label{prop:correctness} When all users are honest, the sequence of iterates $(\widetilde{X}(t))_{t\geq 0}$ generated by \gopa satisfies $\lim_{t\rightarrow+\infty}\mathbb{E}[\widetilde{X}(t)] = X^{avg}\mathbf{1}$.
    Furthermore, the $\tau$-averaging time of Algorithm~\ref{alg:gossip} is:
    \begin{equation}
    \label{eq:tave_gopa}
    t(\tau)=3\log\Big(\frac{2B_\delta(d_{max}+3)}{\tau B_X}\Big)\Big/\log(1/C_G),
    \end{equation}
    where $C_G = 1 - \lambda_2(L)/|E|\in(0,1)$ with $\lambda_2(L)$ the second smallest eigenvalue of the Laplacian matrix $L$ \citep{random_gossip,Colin2015a}, $d_{max} = \max_u d_u$ is the maximum degree and $B_X, B_\delta$ are upper bounds for the absolute value of private values and noise terms respectively.\footnote{We use a bounded noise assumption for simplicity. The argument easily extends to boundedness with high probability.}
  \end{proposition}

  Proposition~\ref{prop:correctness} allows to quantify the \emph{worst-case} impact of the randomization phase on the convergence of the averaging phase. The $\tau$-averaging time of \gopa is only increased by a constant additive factor compared to the non-private averaging phase. Importantly, this additive factor has a mild (logarithmic) dependence on $B$ and $d_{max}$. This behavior will be confirmed by numerical experiments in Section~\ref{sec:exp}.

\subsection{Privacy Guarantees}
\label{sec:privacy}

We now study the privacy guarantees of \gopa. We consider that the knowledge $\mathcal{I}$ acquired by the adversary (colluding malicious users) during the execution of the protocol contains the following: (i) the noisy values $\widetilde{X}$ of all users at the end of the randomization phase, (ii) the full network graph (and hence which pairs of honest users exchanged noise), and (iii) for any communication involving a malicious party, the noise value used during this communication. The only unknowns are the private values of honest users, and noise values exchanged between them. Note that (i) implies that the adversary learns the network average.

\begin{remark}
Since we assume that the noisy values $\widetilde{X}=[\widetilde{X}_1,\dots,\widetilde{X}_n]^\top$ are known to the adversary, our privacy guarantees will hold even if these values are publicly released. Note however that computing the average with Algorithm~\ref{alg:gossip} provides additional protection as the adversary will observe only a subset of the noisy values of honest nodes (those who communicate with a malicious user at their first iteration). In fact, inferring whether a received communication is the first made by that user is already challenging due to the asynchronous nature of the algorithm.
\end{remark}

% The following result precisely characterizes the privacy guarantees of \gopa when the prior and noise distributions are Gaussian.
% \footnote{This result may be extended to other distributions, but we leave this for future work.}

% \aurelien{need to say clearly what is the knowledge of the adversary: it knows (i) the noisy values $\widetilde{X}$ of all users at the end of the randomization phase, (ii) the ``noise communication graph'' over all users (who exchanged noise with whom at what time, even between honest users), and (iii) for any communication involving a malicious party, the noise value. We should emphasize that this is a very strong adversary (the only unknown is essentially the noise values exchanged between honest users). (ii) may seem overly strong but it makes sense when we consider the verification part (because published values reveal who communicated with whom). Clarify that the above knowledge explicitly implies that the adversary knows the network average (in fact, averages over potentially smaller subsets of users when the graph of honest users is not connected)}

Our main result exactly quantifies how much variance is left in any private value $X_u$ after the adversary has observed all information in $\mathcal{I}$.

\begin{theorem}[Privacy guarantees]
\label{thm:variance}
  Assume that the prior belief on the private values $X_1,\dots,X_n$ is Gaussian, namely
  $X_u\sim \mathcal{N}(0,\sigma_X^2)$ for all honest users $u\in U^H$, and that the noise variables are also drawn from a Gaussian distribution $\mu = \mathcal{N}(0,\sigma_\delta^2)$. Denote by $\rvunitvect{u}\in\mathbb{R}^n$ the indicator vector of the $u$-th coordinate and let $M = (I + \frac{\sigmaDelta^2}{\sigmaX^2} L^H )^{-1}\in\mathbb{R}^{n\times n}$, where $L^H$ is the Laplacian matrix of the graph $G^H$. Then we have for all honest user $u^H\in U$:
  \begin{equation}
    \label{eq:variance}
    \frac{var(X_u\mid \mathcal{I})}{var(X_u)} = 1-\rvunitvect{u}^\top M \rvunitvect{u}.
  \end{equation}
%   Denoting by $|N^H(u)|$ the number of honest neighbors of $u$, we further have the following upper bound:
%   \begin{equation}
%     \label{eq:upper_var}
%     var(X_u\mid \mathcal{I}) \geq \sigmaX^2
% \left[
% \frac{\sigmaDelta^2 (\left|N^H(u)\right|+1)
% }{
% \sigmaX^2 + \sigmaDelta^2 (\left|N^H(u)\right|+1)}
% \right]
% \frac{\left|N^H(u)\right|}{\left|N^H(u)+1\right|}.
%   \end{equation}
\end{theorem}

We now provide an detailed interpretation of the above result. First note that the proportion of preserved variance only depends on the interactions between honest users, making the guarantees robust to any adversarial values that malicious users may send.
Furthermore, notice that matrix $M$ can be seen as a smoothing operator over the graph $G^H$. Indeed, given some $y\in\mathbb{R}^n$, $My\in\mathbb{R}^n$ is the solution to the following optimization problem:
\begin{equation}
\label{eq:graph_smoothing}
\min_{s\in\mathbb{R}^n} \|s - y\|_2^2 + \alpha s^\top L^Hs,
\end{equation}
where $\alpha=\sigmaDelta^2/\sigmaX^2\geq 0$. This problem is known as \emph{graph smoothing} (also graph regularization, or graph filtering) and has been used in the context of semi-supervised learning \citep[see][]{Zhu2002a,Zhou2003a}, multi-task learning \citep{Evgeniou2004a} and signal processing on graphs \citep{Shuman2013a}, among others. The first term in \eqref{eq:graph_smoothing} encourages solutions that are close to $y$ while the second term is the Laplacian quadratic form $s^\top L^Hs=\sum_{(u,v)\in E^H}(s_u-s_v)^2$ which enforces solutions that are smooth over the graph (the larger $\alpha$, the more smoothing). One can show that $M$ is a doubly stochastic matrix \citep{Segarra2015a}, hence the vector $My$ sums to the same quantity as the original vector $y$.
In our context, we smooth the indicator vector $\rvunitvect{u}$ so $\rvunitvect{u}^\top M \rvunitvect{u}\in[0,1]$: the larger the noise variance $\sigmaDelta^2$ and the more densely connected the graph of honest neighbors, the more ``mass'' is propagated from user $u$ to other nodes, hence the closer to $0$ the value $\rvunitvect{u}^\top M \rvunitvect{u}$ and in turn the more variance of the private value $X_u$ is preserved. 
Note that we recover the expected behavior in the two extreme cases: when $\sigmaDelta^2=0$ we have $\rvunitvect{u}^\top M \rvunitvect{u}=1$ and hence the variance ratio is $0$, while when $\sigmaDelta^2\rightarrow\infty$ we have $\rvunitvect{u}^\top M \rvunitvect{u}\rightarrow 1/|U_H|$ and hence $var(X_u\mid \mathcal{I})/var(X_u)=1-1/|U_H|$. This irreducible variance loss accounts for the fact that the adversary learns from $\mathcal{I}$ the average over the set of honest users (by adding their noisy values and subtracting the noise terms he knows). Since we assume the number of users to be very large, this variance loss can be considered negligible.
We emphasize that in contrast to the lower bounds on the variance of the maximum likelihood estimate obtained by \citet{Mo2014a}, our variance computation \eqref{eq:variance} is exact, interpretable and holds under non-uniform prior beliefs of the adversary.

It is important to note that smoothing occurs over the entire graph $G^H$: by transitivity, all honest users in the network contribute to keeping the private values safe. In other words, \gopa turns the massively distributed setting into an advantage for privacy, as we illustrate numerically in Section~\ref{sec:exp}. Note, still, that one can derive a simple (but often loose) lower bound on the preserved variance which depends only on the local neighborhood.

\begin{proposition}
\label{prop:var_bound}
  Let $u\in U^H$ and denote by $|N^H(u)|$ the number of honest neighbors of $u$. We have:
  \begin{equation*}
    % \label{eq:upper_var}
    var(X_u\mid \mathcal{I}) \geq \sigmaX^2\cdot
\frac{\sigmaDelta^2 (\left|N^H(u)\right|+1)
}{
\sigmaX^2 + \sigmaDelta^2 (\left|N^H(u)\right|+1)}\cdot
\frac{\left|N^H(u)\right|}{\left|N^H(u)+1\right|}.
  \end{equation*}
\end{proposition}

Proposition~\ref{prop:var_bound} shows that the more honest neighbors, the larger the preserved variance. In particular, if $\sigmaDelta^2>0$, we have $var(X_u\mid \mathcal{I}) \rightarrow \sigmaX^2$ as $|N^H(u)| \rightarrow \infty$.

We conclude this subsection with some remarks.

\begin{remark}[Composition]
\label{rem:composition}
As can be seen from inspecting the proof of Theorem~\ref{thm:variance}, $P(X_u \mid \mathcal{I})$ is a Gaussian distribution. This makes our analysis applicable to the setting where \gopa is run several times with the same private value $X_u$ as input, for instance within an iterative machine learning algorithm. We can easily keep track of the preserved variance by recursively applying Theorem~\ref{thm:variance}.
\end{remark}

\begin{remark}[$G^H$ not connected]
\label{rem:notconnected}
The results above still hold when Assumption~\ref{assump:GH} is not satisfied. In this case, the smoothing occurs separately within each connected component of $G^H$, and the irreducible variance loss is ruled by the size of the connected component of $G^H$ that the user belongs to (instead of the total number of honest users $|U_H|$).
%Clearly, the adversary learns the value of potential isolated users (who have only malicious neighbors).
\end{remark}

\begin{remark}[Drop out]
  The use of centered Gaussian noise with bounded variance effectively limits the impact of some users dropping out during the randomization phase. In particular, any residual noise term has expected value $0$, and can be bounded with high probability. Alternatively, one may ask each impacted user to remove the noise exchanged with users who have dropped (thereby ensuring exact convergence at the expense of reducing privacy guarantees).
\end{remark}

% \begin{theorem}[Privacy guarantees]
%   \label{thm:privacy}
%   \aurelien{to update: give exact expression for $\epsilon$ with Laplacian, and lower/upper bound on that expression? (case of infinite noise variance and case where neighbors only communicate with the node of interest)}
%   Assume that the prior belief on the private values $X_1,\dots,X_n$ is Gaussian, namely
%   $X_u\sim \mathcal{N}(0,\sigma_X^2)$ for all $u\in U$, and that the noise variables $\delta_.$ are
%   drawn from a probability distribution 
%   which is a
%   $B$-bounded Gaussian, i.e., if $|x|\le B$ then
%   $P^{prior}(\Delta=x) = (1/\sqrt{2\pi\sigma_\Delta})\exp(-x^2/2\sigma_\Delta^2)/erf(B)$ else $P^{prior}(\Delta=x)=0$. 
%   Assume every honest user communicates with at least $H$
%   other honest users.
%   Let $P^{post}$ be the posterior distribution after some party has observed
%   all information collected by malicious users.  Then for all $u\in U$,  $e^{-\epsilon}P^{prior}(X_u=z) \le P^{post}(X_u=z) \le e^\epsilon P^{prior}(X_u=z)$, where $\epsilon=B^2 \sigma_X/\sqrt{\sigma_X^2+H\sigma_\Delta^2}$.
% \end{theorem}

\subsection{Robust Strategies for Network Construction}
\label{sec:graphs}

We have seen above that the convergence rate and most importantly the privacy guarantees of GOPA crucially depend on the network topology. In particular, it is crucial that the network graph $G=(U, E)$ is constructed in a robust manner to ensure good connectivity and to guarantee that all honest users have many honest neighbors with high probability. In the following, we assume users have an address list for the set of users and can select some of them as their neighbors. Note that the randomization phase can be conveniently executed when constructing the network.

A simple choice of network topology is the complete graph, which is best in terms of privacy (since each honest user has all $|U_H|-1$ other honest users as neighbors) and convergence rate (best connectivity). Yet this is not practical when $n$ is very large: beyond the huge number of pairwise communications needed for the randomization phase, each user also needs to create and maintain $n-1$ secure connections, which is costly \citep{Chan2003a}.

We propose instead a simple randomized procedure to construct a sparse network graph with the desired properties based on \emph{random $k$-out random graphs} \citep{Bollobas2001a}, also known as \emph{random $k$-orientable graphs} \citep{Fenner1982a}.
The idea is to make each (honest) user select $k$ other users uniformly at random from the set of all users. Then, the edge $(u,v)\in E$ is created if $u$ selected $v$ or $v$ selected $u$ (or both). This procedure is the basis of a popular key predistribution scheme used to create secure peer-to-peer communication channels in distributed sensor networks \citep{Chan2003a}. \citet{Fenner1982a} show that for any $k\geq 2$, the probability that the graph is $k$-connected goes to $1$ almost surely.
% \footnote{A graph is said to be $k$-connected if there does not exist a set of $k-1$ users whose removal disconnects the graph, i.e., the vertex connectivity of the graph is at least $k$.}
This provides robustness against malicious users. Note also that the number of honest neighbors of a honest node follows a hypergeometric distribution and is tightly concentrated around its expected value $k(1-f)$, where $f$ is the proportion of malicious users.
It is worth noting that the probability that the graph is connected is actually very close to 1 even for small $n$ and $k$. This is highlighted by the results of \citet{Yagan2013a}, who established a rather tight lower bound on the probability of connectivity of random $k$-out network graphs. For instance, the probability is guaranteed to be larger than $0.999$ for $k=2$ and $n=50$.
The algebraic connectivity $\lambda_2(L)$ is also large in practice for random $k$-out graphs, hence ensuring good convergence as per Proposition~\ref{prop:correctness}. % and can be lower bounded through results on the diameter \citep[see][]{Philips1990a}.

%% file: subfiles/verif.tex
\section{Verification Procedure}
\label{sec:verif}

  \begin{algorithm*}[t]
    \floatname{algorithm}{Algorithm}
    \caption{Verification procedure for the randomization phase of \gopa}
    \begin{algorithmic}[1]
      \STATE{{\textbf{Input:}} each user $u$ has generated its own Paillier encryption scheme and has published:}
      \STATE{\hspace*{.5cm}$\bullet$ Public key $K^{pub}_u$, $\varepsilon_u^P(X_u)$ (before the execution of Algorithm~\ref{alg:rndphase})}
      \STATE{\hspace*{.5cm}$\bullet$ Noise values $\varepsilon_u^P(\delta_{u,v})$ (when exchanging with $v \in N(u)$ during Algorithm~\ref{alg:rndphase})}
      \STATE{\hspace*{.5cm}$\bullet$ $\varepsilon_u^P(\widetilde{X}_u)$, $\varepsilon_u^P(\Delta_u)$ (at the end of Algorithm~\ref{alg:rndphase})}
      % \State{Verify that
      % $\prod_{k=1}^C \varepsilon(\eta^i_k)=g_i^{0}\cdot r_{tot}^i
      % \bmod N^2$.}
      \FORALL{user $u \in \userset$} % \COMMENT{verify coherence of users' publications}
      \STATE{$\varepsilon^{\Delta} \gets \prod_{v \in N(u)}\varepsilon_u^P(\delta_{u,v})$, $\varepsilon^{\widetilde{X}} \gets \varepsilon_u^P(X_u)\cdot\varepsilon_u^P(\Delta_u)$}
      \STATE{Verify that $\varepsilon_u^P(\Delta_u)= \varepsilon^{\Delta}$ and
        $\varepsilon_u^P(\widetilde{X}_u) =  \varepsilon^{\widetilde{X}}$ (if not, add $u$ to cheater list)} 
      \ENDFOR
      \FORALL{user $u \in \userset$} % \COMMENT{verify validity of noise exchanges}
      \STATE{Draw random subset $V_u$ of $N(u)$ with $|V_u| =\ceil{(1-\beta)d_u}$, ask $u$ to publish $\delta_{u,v}$ and $r_{\delta_{u,v}}$ for $v\in V_u$}
      \FORALL{$v \in V_u$}
      \STATE{Ask $v$ to publish $r_{\delta_{v,u}}$, verify that $\varepsilon_v^P(\delta_{v,u}) = \varepsilon_v(-\delta_{u,v})$ and $\varepsilon_u^P(\delta_{u,v}) = \varepsilon_u(\delta_{u,v})$ (if not, add $u,v$ to cheater list)}
      \ENDFOR
      \ENDFOR
    \end{algorithmic}
    \label{alg:verif}
  \end{algorithm*}

We have shown in Section~\ref{sec:conv} that \gopa converges to the appropriate average when users are assumed not to tamper with the protocol by sending incorrect values. In this section, we complement our algorithm with a verification procedure to detect malicious users who try to influence the output of the algorithm (we refer to them as \emph{cheaters}). While it is impossible to force a user to give the ``right'' input to the algorithm (no one can force a person to answer honestly to a survey),\footnote{We can use range proofs \citep{Camenisch2008a} to check that each input value lies in an appropriate interval without revealing anything else about the value.} our goal is to make sure that given the input vector $X$, the protocol will either output $\avgx$ or detect cheaters with high probability.
Our approach is based on asking users to publish some (potentially encrypted) values during the execution of the protocol.\footnote{We assume that users cannot change their published values after publication. This could be enforced by relying on blockchain structures as in Bitcoin transactions \citep{Nakamoto2008a}.} The published information should be publicly accessible so that anyone may verify the validity of the protocol (avoiding the need for a trusted verification entity), but should not threatens the privacy.

In order to allow public verification without compromising privacy, we will rely on the Paillier encryption scheme described in Section~\ref{sec:paillier}. For the purpose of this section, we assume that all quantities (private values and noise terms) are in $\mathbb{Z}_{N}$. This is not a strong restriction since with appropriate $N$ and scaling one can represent real numbers from a given interval with any desired finite precision.
% In order to allow public verification without compromising privacy, we will rely on a partially homomorphic encryption scheme to cipher the values while allowing certain operations to be carried out directly on the cypher text (without decrypting). We use the popular Paillier cryptosystem \citep{paillier1999public,paillier}, which is additive.
% Formally, one generates a public (encryption) key $K^{pub}=(N,g)$, where $N=pq$ for $p\neq q$ two independent large primes and $g\in\mathbb{Z}^*_{N^2}$, and a secret (decryption) key $K^{priv} = \lcm(p-1,q-1)$. A message $m\in\mathbb{Z}^*_N$ can then be encrypted into a cypher text with $\varepsilon(m) = g^m \cdot r^N \bmod N^2$,
% where $r$ is drawn randomly from $\mathbb{Z}^*_N$. With
% knowledge of the secret key, one can recover $m$ from $\varepsilon(m)$.
% Furthermore, we have the following homomorphic property: for
% $m_1, m_2 \in \mathbb{Z}^*_N$, $\varepsilon(m_1 + m_2) = \varepsilon(m_1) \cdot \varepsilon(m_2)$.
% For the purpose of this section, we assume that $X_u\in\mathbb{Z}_{N}$ for all $u\in U$. This is not a strong restriction as one can represent real numbers from a given interval with any desired precision by appropriately choosing the value of $N$ and the scaling.
Each user $u$ generates its own Paillier scheme and publishes encrypted values using its encryption operation $\varepsilon_u(\cdot)$. These published cypher texts may not be truthful (when posted by a malicious user), hence we denote them by the superscript $P$ to distinguish them from a valid encryption of a value (e.g., $\varepsilon^P_u(v)$ is the cypher text published by $u$ for the quantity $v$).

Algorithm~\ref{alg:verif} describes our verification procedure for the randomization phase of \gopa (the averaging phase can be verified in a similar fashion). In a first step, we verify the coherence of the publications of each user $u$, namely that $\Delta_u = \sum_{v \in N(u)} \delta_{u,v}$ and $\widetilde{X}_u = X_u + \Delta_u$ are satisfied for the published cypher texts of these quantities. To allow reliable equality checks between cipher texts, some extra care is needed (see supplementary material for details).
The second step is to verify that during a noise exchange, the user and his neighbor have indeed used opposite values as noise terms. However, each user has his own encryption scheme so one cannot meaningfully compare cypher texts published by two different users. To address this issue, we ask each user $u$ to publish \emph{in plain text} a random selected fraction $(1-\beta) \in [0,1]$ of his noise values $(\delta_{u,v})_{v \in N(u)}$ (the noise values to reveal are drawn publicly). The following result lower bounds the probability of catching a cheater.

  \begin{proposition}[Verification]
    \label{prop:verif}
    Let $C$ be the number of times a user cheated during the randomization phase. If we apply the verification procedure (Algorithm~\ref{alg:verif}), then the
    probability that a cheater is detected is at least $1-\beta^{2C}$.
  \end{proposition}

  Proposition~\ref{prop:verif} shows that Algorithm~\ref{alg:verif} guards against large-scale cheating: the more cheating, the more likely at least one cheater gets caught. Small scale cheating is less likely to be detected but cannot affect much the final output of the algorithm (as values are bounded with high probability and the number of users is large).
  %\textcolor{red}{missing privacy consequences of publishing a fraction of noise in plain text}
  Of course, publishing a fraction of the noise values in plain text decreases the privacy guarantees: publishing $\delta_{u,v}$ in plain text for $u,v\in U^H$ corresponds to ignoring the associated edge of $G^H$ in our privacy analysis. If the community agrees on $\beta$ in advance, this effect can be easily compensated by constructing a network graph of larger degree, see Section~\ref{sec:graphs}.

%% file: subfiles/experiments.tex
% !TEX root = ../main_supp.tex

\section{Numerical Experiments}
\label{sec:exp}

\begin{figure}[t]
    \centering
    \if\arxiv1
    \subfigure[Relative error w.r.t. \# iter.]{\label{fig:conv1}\includegraphics[width=.35\textwidth]{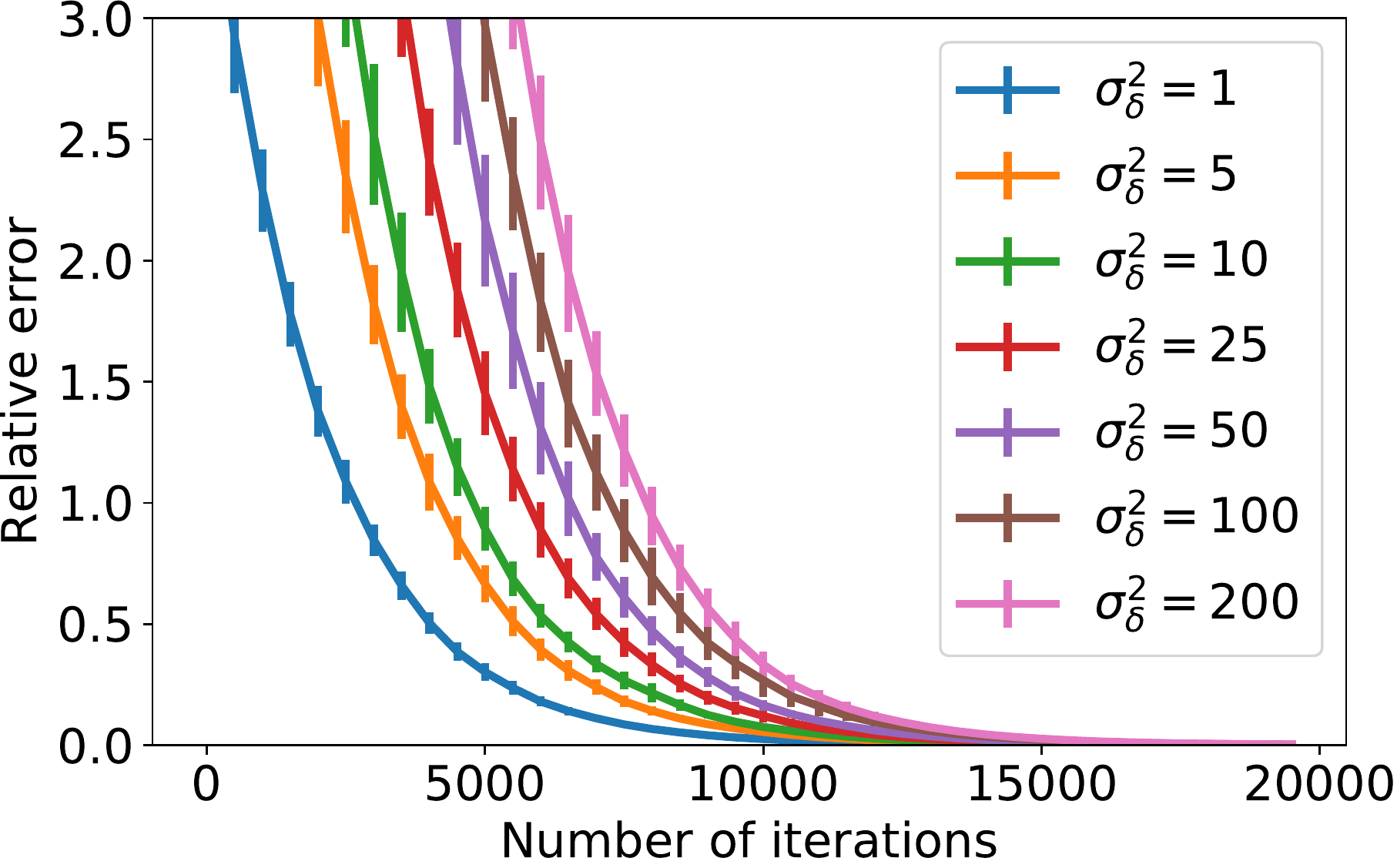}}
    \subfigure[\# iter. to $10^{-2}$ error w.r.t. $\sigma^2_\delta$]{\label{fig:conv2}\includegraphics[width=.35\textwidth]{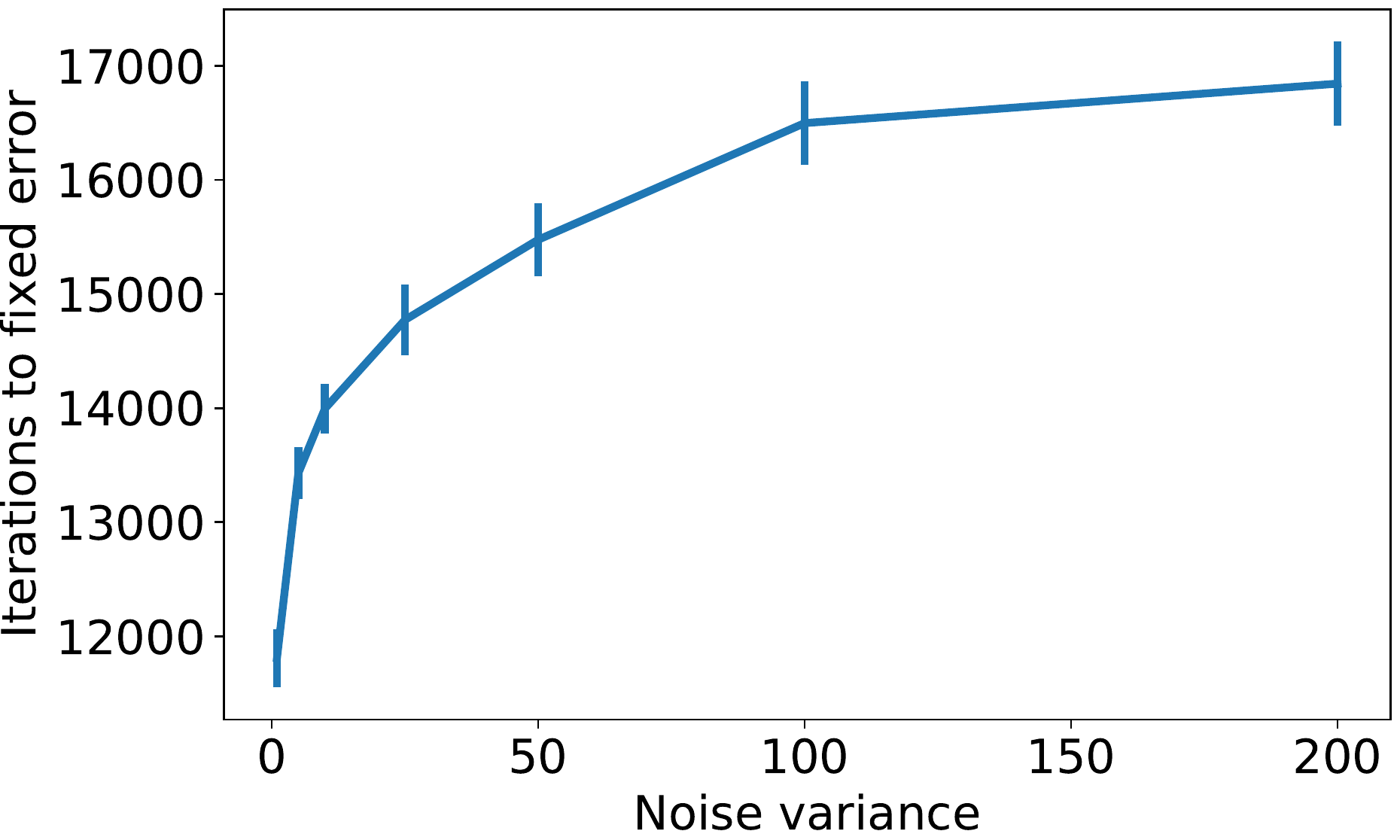}}
    \else
    \subfigure[Relative error w.r.t. \# iter.]{\label{fig:conv1}\includegraphics[width=.235\textwidth]{figures/conv1}}
    \subfigure[\# iter. to $10^{-2}$ error w.r.t. $\sigma^2_\delta$]{\label{fig:conv2}\includegraphics[width=.24\textwidth]{figures/conv2}}
    \fi
    \caption{Impact of the noise variance on the convergence of \gopa for a random $k$-out graph with $n=1000$ and $k=10$, with mean and standard deviation over 10 random runs.}
    \label{fig:conv}
\end{figure}

\begin{figure}[t]
    \centering
    \if\arxiv1
    \subfigure[10\% of malicious nodes]{\label{fig:var1}\includegraphics[width=.35\textwidth]{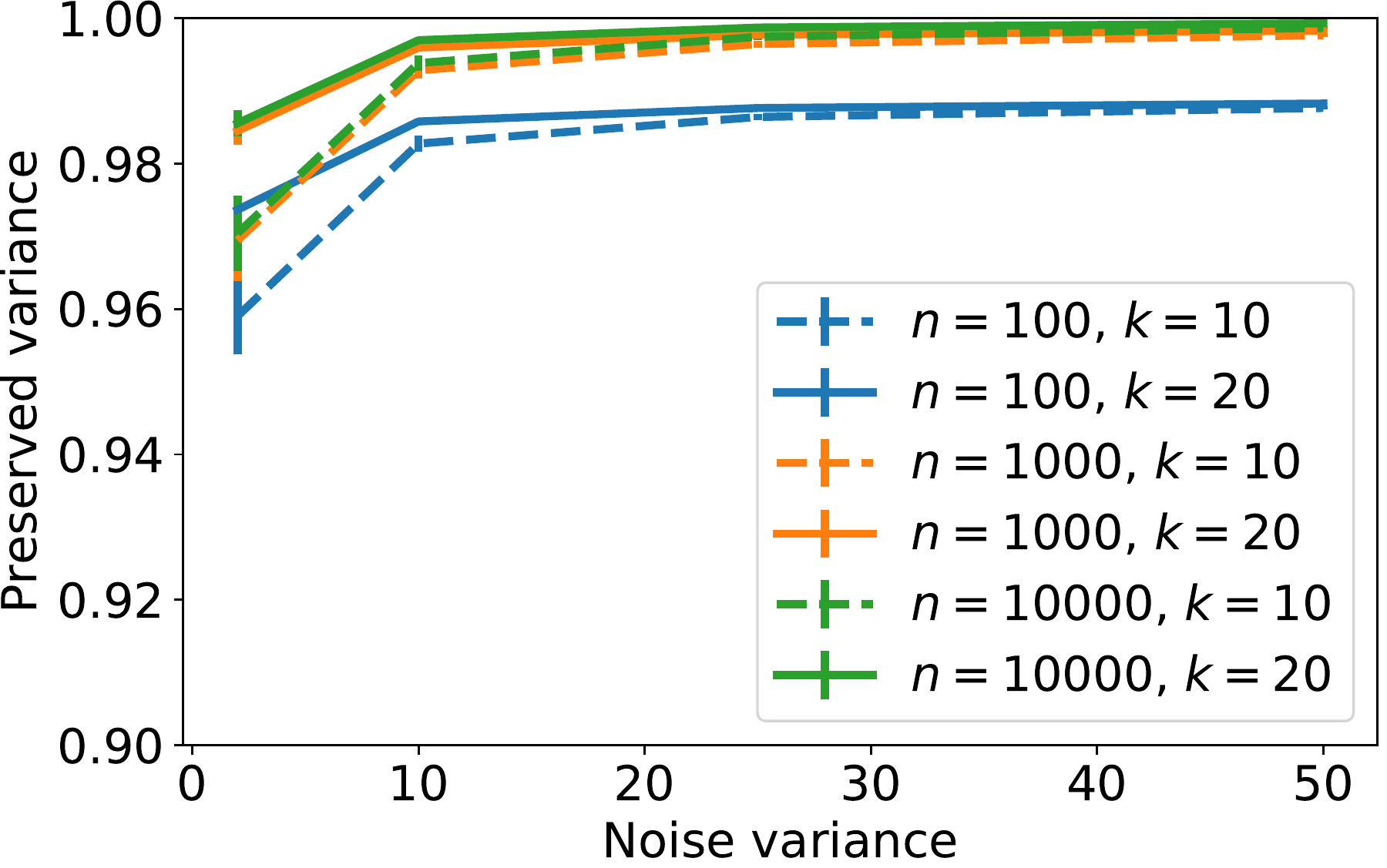}}
    \subfigure[50\% of malicious nodes]{\label{fig:var2}\includegraphics[width=.35\textwidth]{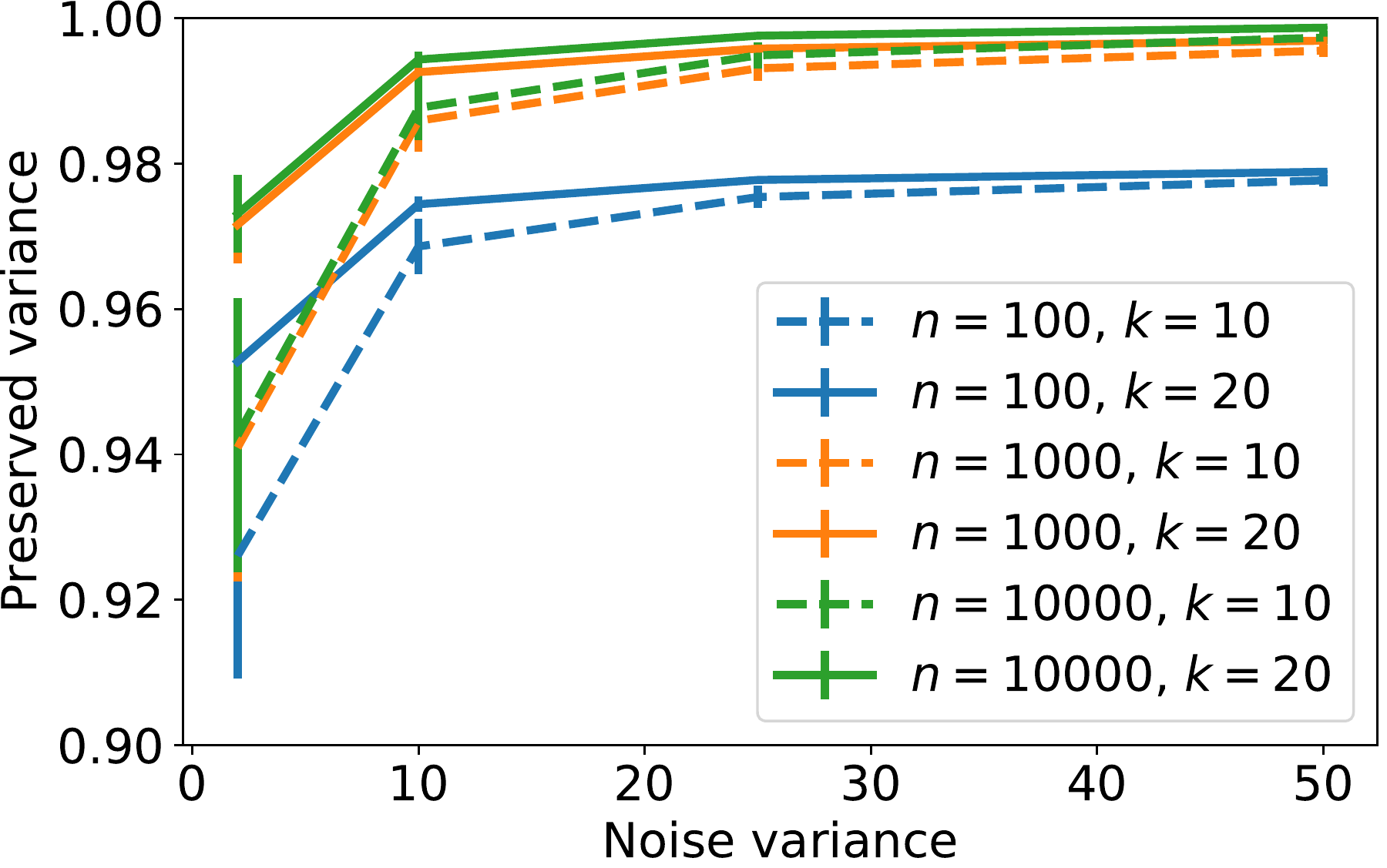}}
    \else
    \subfigure[10\% of malicious nodes]{\label{fig:var1}\includegraphics[width=.23\textwidth]{figures/var_m01}}
    \subfigure[50\% of malicious nodes]{\label{fig:var2}\includegraphics[width=.23\textwidth]{figures/var_m05}}
    \fi
    \caption{Preserved data variance (mean and standard deviation across users) w.r.t. noise variance for several random $k$-out topologies and two proportions of malicious nodes.}
    \label{fig:var}
\end{figure}

In this section, we run some simulations to illustrate the practical behavior of \gopa.
In particular, we study two aspects: the impact of the noise variance on the convergence and on the proportion of preserved data variance, and the influence of network topology.
In all experiments, the private values are drawn from the normal distribution $\mathcal{N}(0,1)$, and the network is a random $k$-out graph (see Section~\ref{sec:graphs}).

Figure~\ref{fig:conv} illustrates the impact of the noise variance on the convergence of the averaging phase of \gopa, for $n=1000$ users and $k=10$. We see on Figure~\ref{fig:conv1} that larger $\sigma_\delta^2$ have a mild effect on the convergence rate. Figure~\ref{fig:conv2} confirms that the number of iterations $t$ needed to reach a fixed error $\|\widetilde{X}(t)-X^{avg}\mathbf{1}\|/\|X\|$ is logarithmic in $\sigma_\delta^2$, as shown by our analysis (Proposition~\ref{prop:correctness}).
Figure~\ref{fig:var} shows the proportion of preserved data variance for several topologies and proportions of malicious nodes in the network. Recall that in random $k$-out graphs with fixed $k$, the average degree of each user is roughly equal to $2k$ and hence remains constant with $n$. The results clearly illustrate one of our key result: beyond the noise variance and the number of honest neighbors, the \emph{total} number of honest users has a strong influence on how much variance is preserved. The more users, the more smoothing on the graph and hence the more privacy, without any additional cost in terms of memory or computation for each individual user since the number of neighbors to interact with remains constant.
This effect is even more striking when the proportion of malicious users is large, as in Figure~\ref{fig:var2}. Remarkably, for $n$ large enough, \gopa can effectively withstand attacks from many malicious users.

%% file: subfiles/conclu.tex
% !TEX root = ../main_supp.tex

\section{Conclusion}
\label{sec:conclu}

We proposed and analyzed a massively distributed protocol to privately compute averages over the values of many users, with robustness to malicious parties.
Our novel privacy guarantees highlight the benefits of the large-scale setting, allowing users to effectively ``hide in the crowd'' by distributing the knowledge of their private value across many other parties.
We believe this idea to be very promising and hope to extend its scope beyond the problem of averaging. In particular, we would like to study how our protocol may be used as a primitive to learn complex machine learning models in a privacy-preserving manner.

% \aurelien{to rework}
% Amongst others, we see two major directions of future work. 
% First, this preliminary version does not
% consider strategies to select a sufficiently large set of users to interact with such that one can be (almost) sure that this set contains enough honest users.
% We also hope to study more closely the consequences of a user dropping out during the execution of the protocol.  We anticipate that one can bound the error on the output by making sure that the sum of noise terms remains sufficiently small.

%% file: subfiles/supplementary.tex
% !TEX root = ../main_supp.tex

\section{Proof of Proposition~\ref{prop:correctness}}

The randomization phase consists in pairs of users adding noise terms which sum to zero. Hence, in the HBC setting, we have $\sum_{u\in\userset}\sum_{v \in N(u)} \delta_{u,v}=0$.  Therefore, the sum of the user values over the network remains unchanged after the randomization phase:
$$\sum_{u\in\userset} \widetilde{X}_u = \sum_{u\in\userset}\left[ X_u + \sum_{v \in N(u)} \delta_{u,v}\right]=\sum_{u\in\userset}X_u.$$
The first claim then follows from the correctness of the averaging procedure of \citet{random_gossip}
used for the averaging phase.

From \citet{random_gossip}, we know that the $\tau$-averaging time of the averaging phase applied to the original (non-noisy) values $X$ is $3\log(1/\tau)/\log(1/C_G)$.
Hence, to achieve the same guarantee \eqref{eq:tave}, we need to run the algorithm for at least
\begin{equation}
\label{eq:newkappa}
t(\tau)=\frac{3\log\left(\frac{1}{\tau}\frac{\|\widetilde{X}\|}{\|X\|}\right)}{\log(1/C_G)}
\end{equation}
iterations. Due to the bound on the noise values and the amount of noise exchanges a user can make, we have for any  $u\in\userset$:
$$B_X-B_\delta(d_{max}+3)\le |\widetilde{X}_u|\le B_X+B_\delta(d_{max}+3),$$
and hence
$$\frac{\|\widetilde{X}\|}{\|X\|} \leq \frac{B_\delta(d_{max}+3)}{B_X}.$$
We get the second claim by plugging this inequality into \eqref{eq:newkappa}.\qed

\section{Proof of Theorem~\ref{thm:variance}}

We first introduce some auxiliary notations. Let $\honestordedgeset=\{(u,v)\in\honestedgeset \mid u<u'\}$ be the ordered edges between honest users.
Let $\allrvs^{(X)}=(X_u)_{u\in\honestuserset}$ be the vector of the private values of all honest users and $\allrvs^{(\delta)}=(\delta_{u,v})_{(u,v)\in\honestordedgeset}$ be the vector of all values exchanged by these users. Let 
$$\allrvs = (\allrvs^{(X)},\allrvs^{(\delta)})\in\mathbb{R}^{|\honestuserset|+|\honestordedgeset|}$$
be the concatenation of these two vectors.
We will index vectors and matrices with elements of $\allrvs$. To emphasize that
we refer to the random variable rather than its value, we will use square brackets.
For $u,v\in\honestuserset$, we define $\usign{u}{v}=1$ and $\smalluser{u}{v}=(u,v)$ if $u<v$, and
  $\usign{u}{v}=-1$ and $\smalluser{u}{v}=(v,u)$ if $u\ge v$.

We formalize the knowledge $\mathcal{I}$ of the adversary by a set of linear equations, specifying the constraints the elements of $\allrvs$ must satisfy according to its knowledge:
\begin{equation}
\label{eq:rv.constr}
\begin{array}{ll}
  X_u+\sum_{v\in \honestneighborset(u)} \usign{u}{v} \delta_{\smalluser{u}{v}} = X^H_u, &\quad \hbox{for all } u\in\honestuserset,\\
\end{array}
\end{equation}

where $X_u^H = \widetilde{X}_u - \sum_{v\in N(U) \setminus \honestneighborset(u)} \delta_{u,v}$ is the noisy value of user $X_u$ minus the sum of all noise terms exchanged with malicious users.

Let $A$ be the matrix representing the above set of linear equations, and $b$ the vector representing the right hand side, so that we can write \eqref{eq:rv.constr} as $A\allrvs=b$.
In particular, $A$ is a sparse $|U_H|\times (|U_H|+|E^<_H|)$ matrix whose elements are zero except for $\forall u\in\honestuserset:A_{u,[X_u]}=1$ and $\forall (u,v)\in\honestordedgeset: A_{u,[\delta_{u,v}]}=1$ and $A_{v,[\delta_{u,v}]}=-1$.

The matrix $A$ has full rank and can be written as
$A=\left[I_{|\honestuserset|}\ B^H\right]$ where $I_{|\honestuserset|}$ is the $|\honestuserset|\times |\honestuserset|$ identity matrix and $B^H$ is the oriented incidence matrix of the graph $G^H$ (where the direction of edges is given by $\honestordedgeset$).
Let $C$ be a full rank $|\honestordedgeset|\times (|\honestuserset|+|\honestordedgeset|)$ matrix such that $AC^\top = 0$.
Let $T=\left[\begin{array}{c}A\\C \end{array}\right]$. As $A$ and $C$ are non-singular and orthogonal to each other, $T$ is also non-singular.
Let $\Sigma$ be a diagonal matrix with
$\forall u\in\honestuserset, \Sigma_{[X_u],[X_u]} = \sigmaX^2$ and
$\forall (u,v)\in\honestordedgeset, \Sigma_{[\delta_{u,v},\delta_{u,v}]} = \sigmaDelta^2$.  We know that $var(\allrvs)=\Sigma$.
$T\allrvs$ is a linear transformation of a multivariate Gaussian, and hence a multivariate Gaussian itself. In particular, it holds that $var(T\allrvs) = T\Sigma T^\top$.

We now consider the random vector $C\allrvs$ conditioned on $A\allrvs=b$.  This again gives a Gaussian distribution, with
\begin{eqnarray*}
  var(C\allrvs|A\allrvs=b) &=& var(C\allrvs) - cov(C\allrvs,A\allrvs) (var(A\allrvs,A\allrvs))^{-1} cov(A\allrvs,C\allrvs) \\
  &=& (C\Sigma C^\top) - (C\Sigma A^\top) (A\Sigma A^\top)^{-1} (A\Sigma C^\top)  \\
  &=& C(\Sigma - \Sigma A^\top (A\Sigma A^\top)^{-1} A\Sigma) C^\top.
\end{eqnarray*}

Let $u\in U^H$ some honest user. We now focus on the variance of the private value $X_u$ conditioned on the information obtained by the adversary:
\begin{eqnarray}
  var(X_u\mid A\allrvs=b)
  &=&
  var(\rvunitvect{u}^\top\allrvs|A\allrvs=b)\nonumber
  \\
  &=&
  \rvunitvect{u}^\top C^\dagger C (\Sigma - \Sigma A^\top (A\Sigma A^\top)^{-1} A\Sigma) C^\top (C^\top)^\dagger \rvunitvect{u}\nonumber \\
  &=& \sigma_X^2 - \rvunitvect{u}^\top\sigma_X^2 A^\top (A\Sigma A^\top)^{-1} A
 \sigma_X^2 \rvunitvect{u}\nonumber \\
  &=& \sigma_X^2 - \sigmaX^4\rvunitvect{u}^\top A^\top (A\Sigma A^\top)^{-1} A
 \rvunitvect{u}\nonumber \\
 &=& \sigma_X^2 - \sigmaX^4\rvunitvect{u}^\top
\left[\begin{array}{c} I \\ {B^H}^\top\end{array}\right]
(\left[\begin{array}{cc} I & B^H\end{array}\right]
\Sigma \left[\begin{array}{c} I \\ {B^H}^\top\end{array}\right]
)^{-1} \left[\begin{array}{cc} I & B^H \end{array}\right]
 \rvunitvect{u}\nonumber \\
 &=& \sigma_X^2 - \sigmaX^4\rvunitvect{u}^\top
\left[\begin{array}{c} I \\ {B^H}^\top\end{array}\right]
(\sigmaX^2 I + \sigmaDelta^2B^H{B^H}^\top
)^{-1} \left[\begin{array}{cc} I & B^H \end{array}\right]
\rvunitvect{u}\nonumber \\
 &=& \sigma_X^2 - \sigmaX^4\rvunitvect{u}^\top
(\sigmaX^2 I + \sigmaDelta^2B^H{B^H}^\top )^{-1}
\rvunitvect{u}\nonumber \\
 &=& \sigma_X^2 - \sigmaX^4\rvunitvect{u}^\top
(\sigmaX^2 I + \sigmaDelta^2 L^H )^{-1}
\rvunitvect{u}\label{eq:lasteqthm1},
\end{eqnarray}
where $L^H$ Laplacian matrix associated with $G^H$. Finally, for clarity we rewrite
$$\sigma_X^2 - \sigmaX^4\rvunitvect{u}^\top
(\sigmaX^2 I + \sigmaDelta^2 L^H )^{-1}
\rvunitvect{u} = \sigma_X^2 [1-\rvunitvect{u}^\top (I + \frac{\sigmaDelta^2}{\sigmaX^2} L^H )^{-1} \rvunitvect{u}].$$

\section{Proof of Proposition~\ref{prop:var_bound}}

To better understand Theorem~\ref{thm:variance}, we investigate a fictional scenario where the only noise exchanged in the network is between the user of interest $u$ and his neighbors. It is clear that the amount of variance preserved in this scenario gives a lower bound for the general case (where pairs of nodes not involving $u$ also exchange noise). Indeed, noise exchanges involving a malicious user have no effect (they are subtracted away in the linear system \eqref{eq:rv.constr}), while those between honest users can only increase the uncertainty for the adversary.

Without loss of generality, assume $u=1$. The Laplacian matrix $L^H$ in Theorem~\ref{thm:variance} (ignoring the nodes which did not exchange noise) is given by
\[
L=\left[\begin{matrix}
|\honestneighborset(u)| & -1 & -1 & \ldots & -1 \\
-1                 &  1 &  0 & \ldots &  0 \\
-1                 &  0 &  1 & \ldots &  0 \\
\vdots             &\vdots&\vdots&        &\vdots\\
-1                 &  0 &  0 & \ldots &  1 
\end{matrix}\right].
\]

Let $V$ be a unitary $(|\honestneighborset(u)|+1)\times(|\honestneighborset(u)|+1)$ matrix for which
\begin{eqnarray*}
V_{:,1}^\top&=& \frac{1}{\sqrt{|\honestneighborset(u)+1|^2-|\honestneighborset(u)|-1}} \left(|\honestneighborset(u)|, -1, \ldots, -1\right), \\
V_{:,2}^\top &=& \frac{1}{\sqrt{|\honestneighborset(u)|+1}} \left(1, \ldots, 1\right).
\end{eqnarray*}
We have \[L=V\diag\left(|\honestneighborset(u)|+1,0,1, \ldots, 1\right) V^\top .\]
Indeed, we can easily verify that $LV_{:,1}=(|\honestneighborset(u)|+1)V_{:,1}$ and $LV_{:,2}=0$.  For every vector $V_{:,i}$ with $i\ge 3$, 
we can check that $V_{1,1}^2+V_{1,2}^2=1$ and hence $V_{1,i}=0$.  It follows that for any $j\ge 2$, $L_{j,:}V_{:,i}=V_{j,i}$.  We know that $V_{:,i}$ is orthogonal to $V_{:,2}$ and hence $\sum_{j=2}^{|\honestneighborset(u)|+1} V_{j,i}=0$.  Then, $L_{1,:}V_{:,i} =
-\sum_{j=2}^{|\honestneighborset(u)|+1} V_{j,i} = 0 = V_{1,i}$. Therefore, all vectors orthogonal to $V_{:,1}$ and $V_{:,2}$ are eigenvectors with eigenvalue $1$.

We can now rewrite the matrix inverse in \eqref{eq:lasteqthm1} from the proof of Theorem~\ref{thm:variance} as:
\begin{eqnarray*}
\left(\sigmaX^2 I + \sigmaDelta^2 L^H\right)^{-1} 
&=& \left(\sigmaX^2 V V^\top + \sigmaDelta^2 V\diag\left(|\honestneighborset(u)|+1,0,1,\ldots, 1\right) V^\top\right)^{-1} \\
&=& V\left(\diag(\sigmaX^2) + \sigmaDelta^2 \diag\left(|\honestneighborset(u)|+1,0,1,\ldots, 1\right)  \right)^{-1}V^\top \\
&=& V\left(\diag\left(\sigmaX^2 + \sigmaDelta^2 (|\honestneighborset(u)|+1),\sigmaX^2 ,\sigmaX^2 + \sigmaDelta^2 , \ldots, \sigmaX^2 + \sigmaDelta^2 \right)  \right)^{-1}V^\top \\
&=& V\diag\left(\left(\sigmaX^2 + \sigmaDelta^2 (|\honestneighborset(u)|+1)\right)^{-1},\sigmaX^{-2} ,\left(\sigmaX^2 + \sigmaDelta^2\right)^{-1},  \ldots, \left(\sigmaX^2 + \sigmaDelta^2)\right)^{-1} \right) V^\top
\end{eqnarray*}

We are interested in user $u=1$ so we focus on the value
\begin{eqnarray*}
\left((\sigmaX^2 I + \sigmaDelta^2 L)^{-1} \right)_{1,1} 
&=& V_{:,1} 
\diag\left(\left(\sigmaX^2 + \sigmaDelta^2 (|\honestneighborset(u)|+1)\right)^{-1},\sigmaX^{-2} ,\left(\sigmaX^2 + \sigmaDelta^2\right)^{-1} ,  \ldots, \left(\sigmaX^2 + \sigmaDelta^2)\right)^{-1} \right) 
V_{:,1}^\top \\
&=& V_{1,1}^2 \left(\sigmaX^2 + \sigmaDelta^2 (|\honestneighborset(u)|+1)\right)^{-1} + V_{1,2}^2 \sigmaX^{-2} \\
&=& \left(\sigmaX^2 + \sigmaDelta^2 (|\honestneighborset(u)|+1)\right)^{-1}
\frac{1}{(|\honestneighborset(u)|+1)^2-|\honestneighborset(u)|-1} \left(|\honestneighborset(u)|\right)^2 + \sigmaX^{-2}\frac{1}{|\honestneighborset(u)|+1} \\
&=& \left(\sigmaX^2 + \sigmaDelta^2 (|\honestneighborset(u)|+1)\right)^{-1}
\frac{|\honestneighborset(u)|}{|\honestneighborset(u)|+1}  + \sigmaX^{-2}\frac{1}{|\honestneighborset(u)|+1}.
\end{eqnarray*}

Plugging back in \eqref{eq:lasteqthm1}, we finally get:
\begin{eqnarray*}
var(X_u| A\allrvs=b) 
&=& \sigmaX^2 - \sigmaX^4 
\left[
\left(\sigmaX^2 + \sigmaDelta^2 (|\honestneighborset(u)|+1)\right)^{-1}
\frac{|\honestneighborset(u)|}{|\honestneighborset(u)|+1}  + \sigmaX^{-2}\frac{1}{|\honestneighborset(u)|+1}
\right] \\ 
&=& \sigmaX^2 - \sigmaX^4 
\left(\sigmaX^2 + \sigmaDelta^2 (|\honestneighborset(u)|+1)\right)^{-1}
\frac{|\honestneighborset(u)|}{|\honestneighborset(u)|+1}  
 - \sigmaX^2\frac{1}{|\honestneighborset(u)|+1}
 \\ 
&=& 
 \sigmaX^2
\left[
1
- \sigmaX^2
\left(\sigmaX^2 + \sigmaDelta^2 (|\honestneighborset(u)|+1)\right)^{-1}
\right]
\frac{|\honestneighborset(u)|}{|\honestneighborset(u)+1|}  
 \\ 
&=& 
 \sigmaX^2
\left[
\frac{\sigmaDelta^2 |(\honestneighborset(u)|+1)
}{
\sigmaX^2 + \sigmaDelta^2 (|\honestneighborset(u)|+1)}
\right]
\frac{|\honestneighborset(u)|}{|\honestneighborset(u)|+1}.
\end{eqnarray*}

\section{Details on Verification Procedure}

\subsection{Reliable Equality Checks}

To allow reliable equality checks between cipher texts, some extra care is needed. For any $u\in U$ and $v\in N(u)$, let us denote by $r_{\delta_{u,v}}$, $r_{\Delta_u}$, $r_{X_u}$ and $r_{\widetilde{X}_u}$ the random integers generated (and kept private) by user $u$ to respectively encrypt $\delta_{u,v}$, $\Delta_u$, $X_u$ and $\widetilde{X}_u$ using formula \eqref{eq:paillier_enc}. User $u$ should generate these random integers so as to satisfy the following relation (to simplify notations we assume that $N(u)=\{1,\dots,d_u\}$):
\begin{align*}
  r_{X_u}, r_{\Delta_u}(0),r_{\delta_{u,k}} \in \ZNZ, & \quad\forall k \in \{1,\dots,d_u\},\\
  r_{\Delta_u}(k) = r_{X_u}(k-1) \cdot r_{\delta_{u,k}}, & \quad\forall k \in \{1,\dots,d_u\},\\
  r_{\Delta_u} = r_{\Delta_u}(d_u) \bmod N,\\
  r_{\widetilde{X}_u} = r_{X_u} \cdot r_{\Delta_u} \bmod N.
\end{align*}
It is easy to see that the equality checks can then be reliably performed. Crucially, the $r_{\delta_{u,k}}$'s are random so some new randomness is added at each iteration of the above recursion, guaranteeing that the encrypted values of the noise terms and the noise sum $\Delta_u$ are perfectly secure. Note on the other hand that the encryption of the noisy value $\widetilde{X}_u$ may not be perfectly secure, which is not a concern as our privacy guarantees of Section~\ref{sec:privacy} hold under the knowledge of $\widetilde{X}_u$.

\subsection{Proof Sketch of Proposition~\ref{prop:verif}}

  We can assume that users published all the required values (this can be easily verified).
  The first part of Algorithm~\ref{alg:verif} verifies the coherence of the publications, namely that for all $u$, $\widetilde{X}_u = X_u + \sum_{v \in N(u)} \delta_{u,v}$ is satisfied for the published encrypted version of these quantities. If the coherence test passes, then the only way for two malicious users $u$ and $v$ to influence the final average is by adding noise terms such that $\delta_{u,v} \neq -\delta_{v,u}$.
  
  Assume $u$ and $v$ indeed cheated. The probability that neither $u$ or $v$ is forced to publish the corresponding noise value is $\beta^2$. Assume either one of them, say $u$, is forced to publish his value $\delta_{u,v}$ in plain text. We can use it to check whether the cypher texts published by both $u$ and $v$ are indeed encrypted versions of $\delta_{u,v}$ and $-\delta_{u,v}$ respectively. If the verification fails, then one of them (or both) cheated. Otherwise, this noise exchange did not affect the average.
  We have shown that each time a user cheats, this is detected with probability at least $1-\beta^2$. Then if $C$ is the number of times that some user cheated, the probability of not catching any cheater is $\beta^{2C}$, and the result follows.